\documentclass{svproc}
\usepackage{url}
\usepackage{setspace}
\usepackage{lineno,hyperref}
\usepackage{graphicx}
\usepackage{amsmath}
\usepackage{float}
\usepackage{caption}
\usepackage{subfigure}
\usepackage{algorithm}
\usepackage{booktabs}
\usepackage{algpseudocode}
\usepackage{amsmath}
\usepackage{float}

\begin{document}
\mainmatter              
\title{An improved nonlinear FastEuler AHRS estimation based on the SVDCKF algorithm }
\titlerunning{Nonlinear AHRS FastEuler-SVDCKF Estimation }  
%
\author{Yue Yang \and Xiaoxiong Liu \and
Weiguo Zhang \and Xuhang Liu \and Yicong Guo}
\authorrunning{Yue Yang et al.} 
%
%
\institute{School of Automation, Northwestern Polytechnical University,China,\\
\email{yangyue@mail.nwpu.edu.cn}}

\maketitle              

\begin{abstract}
In this paper, we present a Singular Value Decomposition Cubature Kalman Filter(SVDCKF) fusion algorithm based on the improved nonlinear FastEuler Attitude and Heading Reference and System(AHRS) estimation model for small-UAV attitude. The contributions of this work are the derivation of the low-cost IMU/MAG integrated AHRS model combined with the quaternion attitude determination, and use the FastEuler to correct the gyroscope attitude update, which can increase the real-time solution. In addition, the SVDCKF algorithm is fused the various raw sensors data in order to improve the filter accuracy compared with the CKF. The simulation and experiment results demonstrate the proposed algorithm has the more excellent attitude solution accuracy  compared with the CKF in the low and high dynamic flight conditions.
\end{abstract}

\section{Introduction}
The AHRS\cite{Gebre:Demoz:Roger:Powel,Yadav:Nagesh:Roger:Chris} is the important parts of the autonomous unmanned flight systems(AUFS), and the AHRS is composed of the gyroscope, accelerometer, magnetometer and the microprocessor, and using the information fusion algorithm to calculate the attitude without the assistance of other sensors data. In recent years, the current research on the AHRS algorithm mainly includes the CF, gradient descent filter and nonlinear Kalman Filter so on. The main focus of this paper is the study of the accurate and reliable AHRS algorithm for the small-UAV with the nonlinear Kalman Filter.

The Kalman Filter\cite{Welch:Greg:Gary} is the optimal estimation algorithm that fuses the multiple sensors data with noises in the actual environment, and often used as an optimal estimator in the Gaussian distribution noise systems. However, the Kalman Filter is generally used in the linear systems. The AHRS model has nonlinearity in the practical applications. Thus, the nonlinear Kalman Filter\cite{wang} has been widely studied by the scholars around the world. Aimed at the AHRS of UAV, Song Yu\cite{song:Weng} proposed a quaternion EKF algorithm and designed adaptive filter to correct the measurement noise covariance matrix, which not only solved the problem of large errors about MEMS devices, but also reduced the influence of random errors for the gyroscope on attitude estimation. Yong Liang\cite{Wu} proposed an improved EKF algorithm to estimate the small helicopters attitude, and taking the bidirectional vector systems as the measurement updating, using the flight sensor data to evaluate the filter algorithm performance. Nevertheless, the EKF is the way of first-order Taylor expansion of the function for the nonlinear model, which exists the rounding errors resulting in the problem of divergence and poor filter accuracy. Subsequently, there are some improved nonlinear kalman fiter methods\cite{ZHAO} which include UKF\cite{wan:Rudolph}, CKF\cite{Arasaratnam:Simon} and PF\cite{Arasaratnam:Sanjeev} so on. Pourtakdoust\cite{Pourtakdoust:Ghanbarpour} proposed the quaternion orientation estimation based on the adaptive Unsecnted Kalman Filter, which can capture the effect of nonlinear AHRS model up to second-order without the need for explicit calculations of the Jacobians matrixs. The CKF uses the cubature sample points to approximate the nonlinear model that can up to the third-order compared to the UKF. Although, the PF is a sequential important sampling filter method based on the bayesian sampling estimation, which has better filter accuracy compared with both UKF and CKF. However, there are some disadvantages of the large calculation quantity and the poor real-time. Additionally, the PF is prone to occur the problem of particle starvation causing the filter divergence.  

In this paper, aimed at the characteristics of AHRS model nonlinearity, the CKF combined with the SVD is designed to obtain the better filter accuracy. The decomposition of the QR is substituted by the SVD in order to solve the problem of non-positive definite of the state covariance $P$. The rest of this paper is organized as follows: In Section \ref{sec1}, the mathematical model of the nonlinear AHRS model based on the quaternion solution and the sensor model is introduced; An approach of this study is proposed about SVDCKF fusion with FastEuler algorithm in Section \ref{sec2}; Then, The results of numerical simulation and experimental analysis in Section \ref{sec3} demonstrate the performance of the proposed algorithm; And conclusion is given in Section \ref{sec4}.
\section{IMU/MAG Integrated AHRS Model}\label{sec1}
\subsection{Gyroscope Error Model}
For the low-cost IMU of UAV AHRS estimation, the bias error is the main factor affecting the output accuracy of the MEMS IMU. Eliminating the bias error can improve the accuracy of the three-axis angular velocity of the gyroscope. Additionally, the random drift error of the gyroscope can be changed slowly with time, and it needs to be modeled and estimated.
\begin{equation}
\pmb{\varepsilon}_{g} = \pmb{\varepsilon}_{0} + \pmb{\varepsilon}_{r} + \pmb{\omega}_{\varepsilon}
\label{equ1}
\end{equation}
Where $\pmb{\varepsilon}_{0}$ is the constant value error, $\pmb{\varepsilon}_{r}$ is the random drift error, and $\pmb{\omega}_{\varepsilon}$ is white noise of the gyroscope. The random drift error can be represented by the Eqaution\eqref{equ2}.

\begin{equation}
\pmb{\varepsilon}_{r} = -\frac{1}{\tau_{g}}\pmb{\varepsilon}_{r} + \pmb{\omega}_{\varepsilon r}
\label{equ2}
\end{equation}
Where $\tau_{g}$ is the first-order Markov correlation time constant, $\pmb{\omega}_{\varepsilon r}$ is the white noise. $\pmb{\varepsilon}_{0}$ is the constant value when the IMU is running, and can be derived by the Equation\eqref{equ3}.

\begin{equation}
\left\{
\begin{aligned}
&\pmb{\varepsilon}_{0} = 0 \\
&\pmb{\varepsilon}_{0} + \pmb{\varepsilon}_{r} = \pmb{\varepsilon}_{r}
\end{aligned}
\right.
\label{equ3}
\end{equation}

The constant error can be taken into account in the random drift error, so the gyroscope error model is as follows.
\begin{equation}
\pmb{\varepsilon}_{g} = \pmb{\varepsilon}_{r} + \pmb{\omega}_{\varepsilon}
\label{equ4}
\end{equation}

\subsection{Attitude Error Model}

The strapdown inertial attitude error vector equation is as follows:
\begin{equation}
\delta\pmb{\varphi} = \delta\pmb{\varphi} \times \pmb{\omega}^{n}_{in} + \delta\pmb{\omega}^{n}_{in} - \pmb{\varepsilon}^{n}_{g}
\label{equ5}
\end{equation}
Where the $\delta\pmb{\varphi}$ is the error of attitude about UAV, $\pmb{\omega}^{n}_{in}$ is the angular velocity of the navigation coordinate system relative to the inertial coordinate system, $\pmb{\varepsilon}^{n}_{g}$ is the gyroscope bias in the navigation coordinate system.

The IMU, which is selected in this paper, is a low-cost MEMS device. Compared with high-precision inertial devices, and some small state quantity changes can be submerged in the measurement noise. At the same time, the state estimation values of the attitude error are also small quantity. The attitude error model in Equation\eqref{equ5} is a bit complicated for studying the embedded sensor attitude estimation algorithm. In order to reduce the calculation complexity and increase the algorithm update frequency, some reasonable simplifications can be made to the attitude error model.

Due to the high update frequency of the attitude estimation algorithm, the rotation of the navigation coordinate system caused the UAV position changes can be ignored; the measurement accuracy of the earth's rotational angular rate relative to the gyroscope is also relatively small, and can be approximately 0. In addition, the small UAV has a flight range of several kilometers to several tens of kilometers, and the influence of the radius of earth on the attitude estimation can be ignored. Therefore, the simplified attitude error model can be obtained from Equations\eqref{equ2},\eqref{equ3},\eqref{equ4}.

\begin{equation}
\delta\pmb{\varphi} = - \pmb{\varepsilon}^{n}_{g} = \pmb{C}^{n}_{b}\pmb{\varepsilon}^{b}_{r} -  \pmb{C}^{n}_{b}\pmb{\omega}_{\varphi}
\label{equ6}
\end{equation}

\subsection{Attitude Integrated Model}

Normally, the attitude estimation system has nonlinear characteristics, so it needs to be modeled and established the nonlinear Gaussian state space model\cite{Costanzit:Riccardo}.

\begin{equation}
\left\{
\begin{aligned}
&x(t)  = f(x(t-1)) + w(t-1)\\
&z(t)  = h(x(t)) + v(t)
\end{aligned}
\right.
\label{equ7}
\end{equation}
Where the $x(t)$ is the state estimation parameters, $f(x(t-1))$ is the nonlinear dynamic function, $w(t-1)$ is the process noise, $z(t)$ is the observe parameters, $h(t)$ is the nonlinear observation function, $v(t)$ is the measuremenat noise. Among them, supposed that $w(t)$ and $v(t)$ are zero mean Gaussian white noise and uncorrelated with each other. According to the Section\ref{sec1}, \ref{sec2}, the attitude error Integrated state euqtion can be obtained.

\subsection{Quaternion attitude determination}

The dynamic response of the angular velocity of the gyroscope is relatively fast during the sampling period, and angular velocity $\pmb{\omega^{b}}$ is often taken as constant value. Additionally, because the gyroscope sampling frequency is relative high and the update interval $T = t_{k+1} - t_{k}$ is small, the equivalent rotation vector $\pmb{\Phi}$ can be considered as a relative small amount. Therefore,the equivalent rotation vector $\pmb{\Phi}$ can be expressed by the angular incrementas $\Delta\pmb{B}$.

\begin{equation}
\pmb{\Phi} =\pmb{\omega}^{b}T = [\omega_{x},\omega_{y},\omega{z}]^{T}T = \pmb{\Delta B} = [\Delta B_{x},\Delta B_{y},\Delta B_{z}]^{T}
\label{equ8}
\end{equation}

\begin{equation}
\Phi = \arrowvert \pmb{\Phi} \arrowvert = \Delta B = \sqrt{\Delta B^{2}_{x} + \Delta B^{2}_{y} + \Delta B^{2}_{z}}
\label{equ9}
\end{equation}

The rotation vector is expressed in unit quaternions\cite{Bart:I.:Yaakov} as follows:
\begin{equation}
\pmb{Q}(T) = [cos\frac{\Delta B}{2},\frac{\Delta B_{x}}{\Delta B}sin\frac{\Delta B}{2},\frac{\Delta B_{y}}{\Delta B}sin\frac{\Delta B}{2},\frac{\Delta B_{z}}{\Delta B}sin\frac{\Delta B}{2}]^{T}
\end{equation}
 
According to the quaternion multiplication rule\cite{Shepperd:Stanley}, the attitude update can be expressed by Equation\eqref{equ11}, where the $\bigotimes$ is the quaternion multiplication symbol.
\begin{equation}
\pmb{Q}(t_{k+1}) = \pmb{Q}(t_{k}) \bigotimes \pmb{Q}(T)
\label{equ11}
\end{equation}

\section{FastEuler-SVDCKF AHRS fusion algorithm}\label{sec2}
\subsection{Fast-Euler angle algorithm}

During the attitude solution process, the accelerometer and magnetometer measurements can correct the attitude. but using directly the three-axis acceleration and magnetic measurement values can increase the calculation burden, and it is not easy to detect when abnormal measurement values occur. 

The small-UAV application scenario about the attitude solution in this paper is near-ground navigation, and the flight speed is pretty low. Thus, the Fast-Euler angle algorithm is proposed, which calculate the three-axis accelerometer and magnetic values into the attitude angle, which can be view as the observation values. Consequently, the counts of observation is decreased from the 6 to 3, and increase the filter real-time.
 
\begin{algorithm}[htb]
	\caption{ Fast-Euler angle algorithm.}
	\label{alg:Framwork}
	\begin{algorithmic}[1]
		\Require
		$accel = [a_{x},a_{y},a_{z}]$, $mag = [m_{x},m_{y},m_{z}]$	
		\Ensure
		$\phi_{a}$, $\theta_{a}$, $\psi_{m}$\\	
		\pmb{Step 1}: calculate the roll and pitch observation values 
		\If{$accel \ne 0 $ and $\arrowvert \sqrt{a^{2}_{x} + a^{2}_{y} + a^{2}_{z} } - g \arrowvert \le \alpha $} \\
		$\phi_{a} = atan2(-a_{y},-a_{z})$ \\
		$\theta_{a} = atan2(a_{x},-a_{z})$ 
		\EndIf \\
		\pmb{Step 2}: calculate the yaw observation value 
		\If{$m \ne 0$} \\
		$hx = m_{x}cos\theta_{a} + m_{y}sin\theta_{a}sin\phi_{a} + m_{z}sin\theta_{a}cos\phi_{a}$\\
		$hy = m_{y}cos\phi_{a} - m_{z}sin\phi_{a}$ \\
		$\psi_{m} = atan2(-hy,hx)$ 
		\EndIf \\
		\pmb{Step 3}: judge yaw heading direction
		\If{$\psi_{m} \le 0 $} \\
		$\psi_{m} = \psi_{m} + 2\pi$
		\EndIf		
	\end{algorithmic}
\end{algorithm}|

\subsection{The SVDCKF Sensor Fusion algorithm}

The CKF can provide a systematic solution for high-dimensional nonlinear problem that includes the nonlinear AHRS calculation, it has at least the third-order Taylor series approximation for nonlinear functions compared with the EKF and UKF in the solution accuracy, and it uses the Cholesky decomposition\cite{Higham:Nicholas} of state covariance matrix $\pmb{P}$, $\pmb{P} = \pmb{U}^{T}\pmb{U}$, and $\pmb{U}$ is the triangular matrix. Nevertheless, it can be found several disadvantages during using the Cholesky decomposition: 

1) The Cholesky decomposition defines the matrix $\pmb{P}$ accorded with the properties of the positive definite or symmetric positive definite, and limits the range of the initial value of $\pmb{P}$.

2) The matrix $\pmb{P}$ can become the sparse matrix during the running period of filter algorithm, and destroys the requirements of the Cholesky decomposition. 

This paper employs the Singular Value Decomposition(SVD)\cite{Golub:Gene} to replace the Cholesky decomposition for handling the matrix $\pmb{P}$ which can be expanded to the arbitrary matrix.
\begin{equation}
\begin{aligned}
\pmb{P} = \pmb{U}\pmb{S}\pmb{V}^{T}
\end{aligned}
\label{equ12}
\end{equation}
Where $\pmb{P}$ is the dimensions of $m\times m$ arbitrary matrix, $\pmb{U}$ and $\pmb{V}$ are the unit orthogonal matrix, respectively. the $\pmb{S}$ is the diagonal matrix that has the elements called the singular valuer, and $\pmb{U} \in R^{m \times m}$, $\pmb{S} \in R^{m \times n}$ and $\pmb{V} \in R^{n \times n}$, respectively.
\begin{equation}
\begin{aligned}
\pmb{S} =
\begin{bmatrix}
s_{1} & 0 &0 &0 &0 \\
0 & s_{2} &0 &0 &0 \\
0 & 0 & s_{3} &0 &0 \\
0 & 0 & 0& \ddots&0 \\
0 & 0 & 0& 0& s_{n} \\
\end{bmatrix}
\end{aligned}
\label{equ13}
\end{equation}

Thus, the FastEuler-SVDCKF algorithm is describled in Fig.\ref{fig1}, this paper summarizes the SVDCKF algorithm writing the explicit steps as follows:
 \vspace{-4mm}
 \begin{figure}[htbp]
	 \centerline{\includegraphics[width=4in]{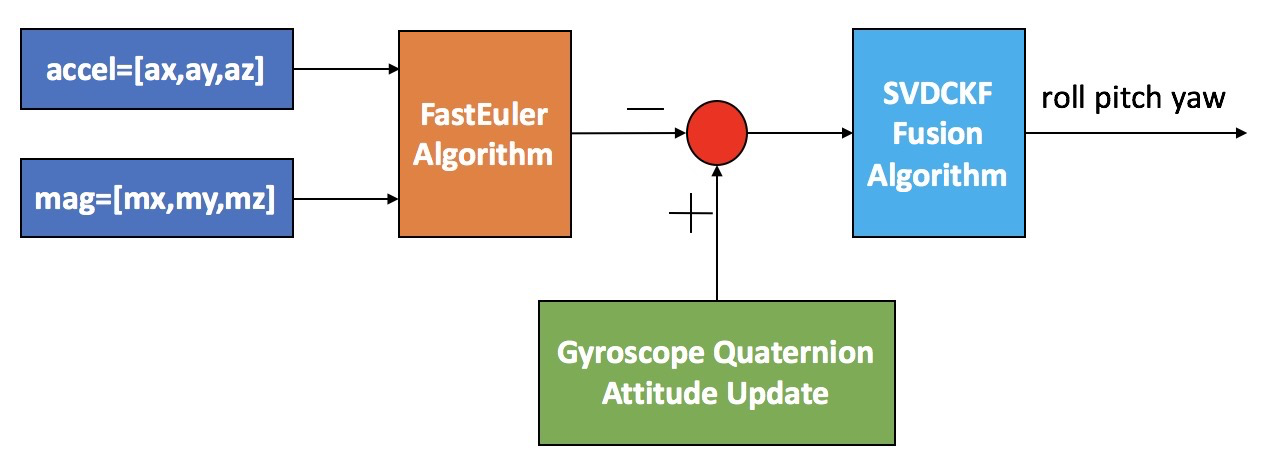}}
	\caption{ The FastEuler-SVDCKF algorithm framework}
	\label{fig1}
 \end{figure}

\pmb{step 1} The setting of initial value $\hat{\pmb{x}}_{0\mid 0 } $ and $\hat{\pmb{P}}_{0\mid 0 }$ in the filter algorithm and calculating the cubature points $\xi_{i} $ and weight $\omega_{i}$ based on the multi-dimension spherical-radial rule.
\begin{equation}
\left\{
\begin{aligned}
& \hat{\pmb{x}}_{0\mid 0 }  = E(\pmb{x}_{0})\\
& \hat{\pmb{P}}_{0\mid 0 }  = E[(\pmb{x}_{0} - \hat{\pmb{x}}_{0\mid 0 })(\pmb{x}_{0} - \hat{\pmb{x}}_{0\mid 0 })^{T}]
\end{aligned}
\right.
\end{equation}
Where$E(.)$ is the expectation.
\begin{equation}
\left\{
\begin{aligned}
&\xi_{i} = \sqrt{\frac{m}{2}}[\pmb{1}]_{i} \\
&\omega_{i} = \frac{1}{m}, i = 1,2,...,m = 2n
\end{aligned}
\right.
\end{equation}
Where n is the dimensions of filter state, m is the number of cubature points, and $[\pmb{1}]_{i}$ is the same points as the following set of points:
\begin{equation*}
\begin{aligned}
\begin{Bmatrix}
\left(
\begin{array}{ccc}
1 \\ 0
\end{array}
\right)
,
\left(
\begin{array}{ccc}
0 \\ 1
\end{array}
\right)
,
\left(
\begin{array}{ccc}
-1 \\ 0
\end{array}
\right)
,
\left(
\begin{array}{ccc}
0 \\ -1
\end{array}
\right)
\end{Bmatrix}_{i}
\end{aligned}
\end{equation*}
\pmb{step 2} State prediction(k = 1,2,3,...):

The covariance matrix $\hat{\pmb{P}}_{k-1\mid k-1}$ is decomposed by the Equation\eqref{equ16}.
\begin{equation}
\begin{aligned}
\hat{\pmb{P}}_{k-1\mid k-1} = \pmb{U}_{k-1\mid k-1}\pmb{S}_{k-1\mid k-1}\pmb{V}^{T}_{k-1\mid k-1}
\end{aligned}
\label{equ16}
\end{equation}
Where $\hat{\pmb{P}}_{k-1\mid k-1}$ is the symmetric matrix, and so $\pmb{U}_{k-1\mid k-1} = \pmb{V}_{k-1\mid k-1}$.

Evaluate the cubature points$(i = 1,2,...,m)$
\begin{equation}
\begin{aligned}
\pmb{X}_{i,k-1 \mid k-1} = \pmb{U}_{i,k-1 \mid k-1}(\sqrt{\pmb{S}_{k-1\mid k-1}})\xi_{i} + \hat{\pmb{x}}_{k-1 \mid k-1}
\end{aligned}
\end{equation}

Evaluate the propagated cubature points through the nonlinear dynamic function $f(.)$$(i = 1,2,...,m)$
\begin{equation}
\begin{aligned}
\pmb{X}^{\ast}_{i,k \mid k-1} = f(\pmb{X}_{i,k-1 \mid k-1},u_{k-1})
\end{aligned}
\end{equation}

Estimate the predicted state and error covariance
\begin{equation}
\left\{
\begin{aligned}
& \hat{\pmb{x}}_{k\mid k-1} = \frac{1}{m}\sum_{i=1}^{m}\pmb{X}^{\ast}_{i,k \mid k-1} \\
& \hat{\pmb{P}}_{k\mid k-1} = \frac{1}{m}\sum_{i=1}^{m}\pmb{X}^{\ast}_{i,k \mid k-1}\pmb{X}^{\ast T}_{i,k \mid k-1} - \hat{\pmb{x}}_{k\mid k-1}\hat{\pmb{x}}^{T}_{k\mid k-1} + \pmb{Q}_{k-1}
\end{aligned}
\right.
\end{equation}

\pmb{step 3} State correction(k = 1,2,3,...):

The covariance matrix $\hat{\pmb{P}}_{k\mid k-1}$ is decomposed by the Equation\eqref{equ20}.
\begin{equation}
\begin{aligned}
\hat{\pmb{P}}_{k\mid k-1} = \pmb{U}_{k\mid k-1}\pmb{S}_{k\mid k-1}\pmb{V}^{T}_{k\mid k-1}
\end{aligned}
\label{equ20}
\end{equation}

Evaluate the cubature points$(i = 1,2,...,m)$
\begin{equation}
\begin{aligned}
\pmb{X}_{i,k \mid k-1} = \pmb{U}_{i,k\mid k-1}(\sqrt{\pmb{S}_{k\mid k-1}})\xi_{i} + \hat{\pmb{x}}_{k\mid k-1}
\end{aligned}
\end{equation}

Evaluate the propagated cubature points through the nonlinear observation function $h(.)$ $(i = 1,2,...,m)$
\begin{equation}
\begin{aligned}
\pmb{Z}_{i,k \mid k-1} = h(\pmb{X}_{i,k \mid k-1},u_{k})
\end{aligned}
\end{equation}

Estimate the predicted measurement, the innovation covariance matrix and the cross-covariance matrix
\begin{equation}
\left\{
\begin{aligned}
& \hat{\pmb{z}}_{k\mid k-1} = \frac{1}{m}\sum_{i=1}^{m}\pmb{Z}^{\ast}_{i,k \mid k-1} \\
& \pmb{P}_{zz,k\mid k-1} = \frac{1}{m}\sum_{i=1}^{m}\pmb{Z}_{i,k \mid k-1}\pmb{Z}^{T}_{i,k \mid k-1} - \hat{\pmb{z}}_{k\mid k-1}\hat{\pmb{z}}^{T}_{k\mid k-1} + \pmb{R}_{k} \\
& \pmb{P}_{xz,k\mid k-1} = \frac{1}{m}\sum_{i=1}^{m}\pmb{X}_{i,k \mid k-1}\pmb{Z}^{T}_{i,k \mid k-1} - \hat{\pmb{x}}_{k\mid k-1}\hat{\pmb{z}}^{T}_{k\mid k-1}
\end{aligned}
\right.
\end{equation}
Estimate the Kalman filter gain
\begin{equation}
\begin{aligned}
\pmb{K}_{k} = \pmb{P}_{xz,k \mid k-1}\pmb{P}^{-1}_{zz,k \mid k-1}
\end{aligned}
\end{equation}
Estimate the updated state and error covariance
\begin{equation}
\left\{
\begin{aligned}
& \hat{\pmb{x}}_{k \mid k-1} = \hat{\pmb{x}}_{k \mid k-1} + \pmb{K}_{k}(\pmb{z}_{k} - \hat{\pmb{z}}_{k \mid k-1}) \\
& \hat{\pmb{P}}_{k \mid k} = \hat{\pmb{P}}_{k \mid k-1} - \pmb{K}_{k}\pmb{P}_{zz,k \mid k-1}\pmb{K}^{T}_{k}
\end{aligned}
\right.
\end{equation}

\section{The Algorithm Simulation}\label{sec3}
\begin{figure}[H]
 	\centerline{\includegraphics[width=3cm,height=2cm]{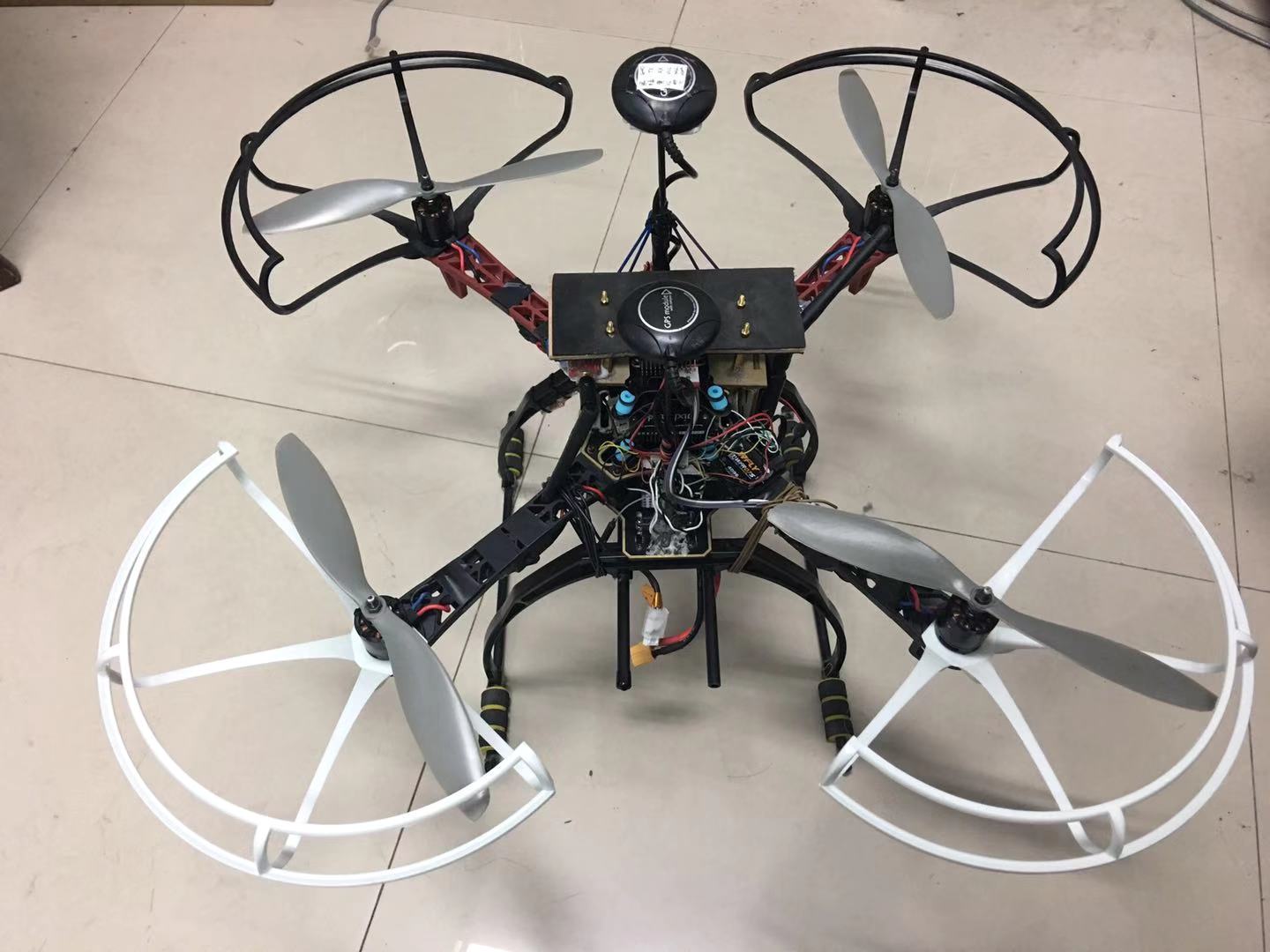} 
 		\quad\qquad 	\includegraphics[width=3cm,height=2cm]{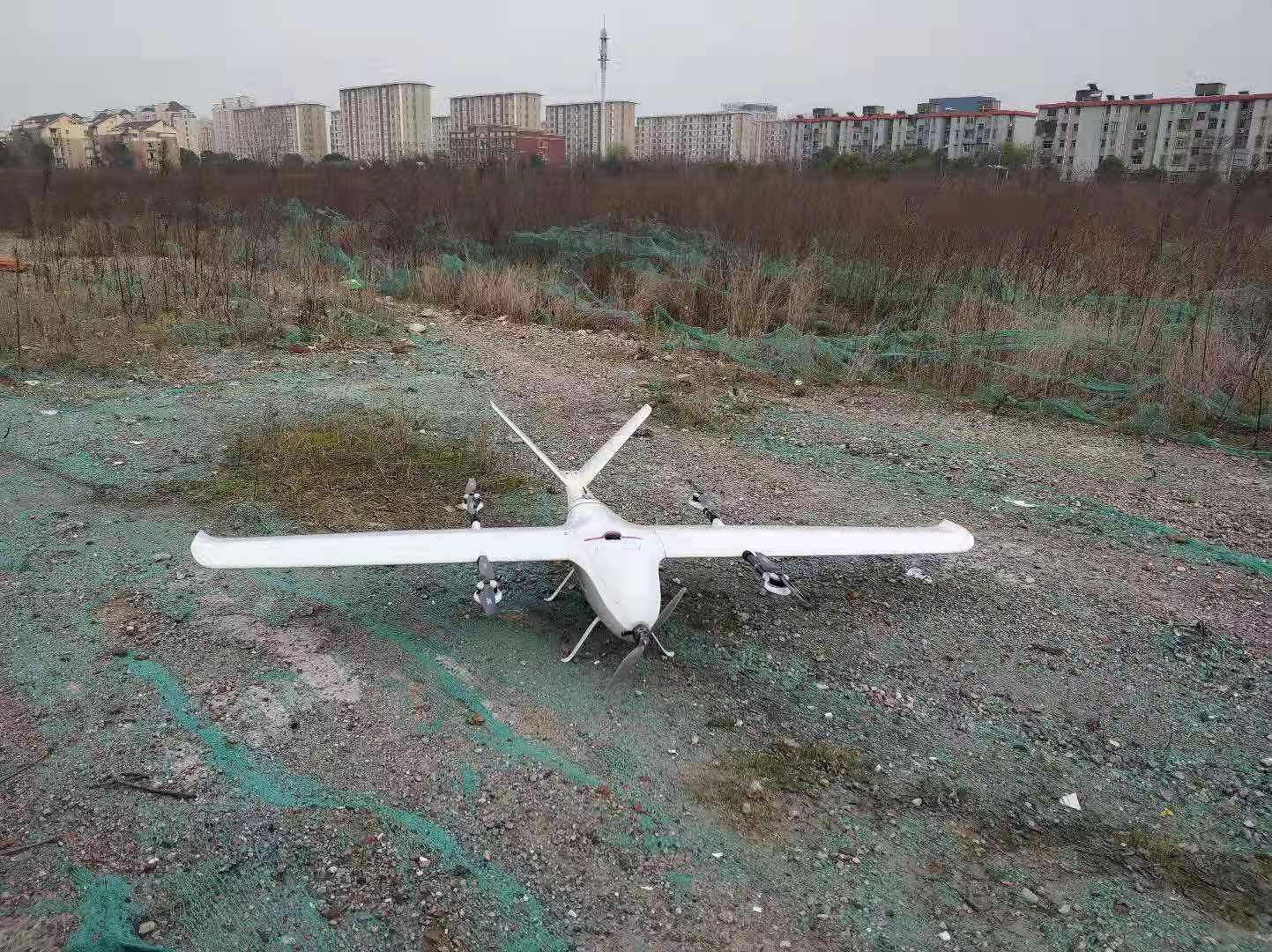}} 
 	\caption{ The data acquisition platform}
 	\label{fig2}
 \end{figure}

 The experimental sensors data of small-UAV from the actual flight environment as shown in Fig.\ref{fig2}, which contains various sensors raw data and attitude truth. Thus this paper uses the proposed algorithm fusing the raw data to calculate the attitude compared with CKF.
  \vspace{-5mm}
 \vspace{-13mm}
 \begin{figure}[htbp]
 	\centerline{\includegraphics[width=6cm,height=4cm]{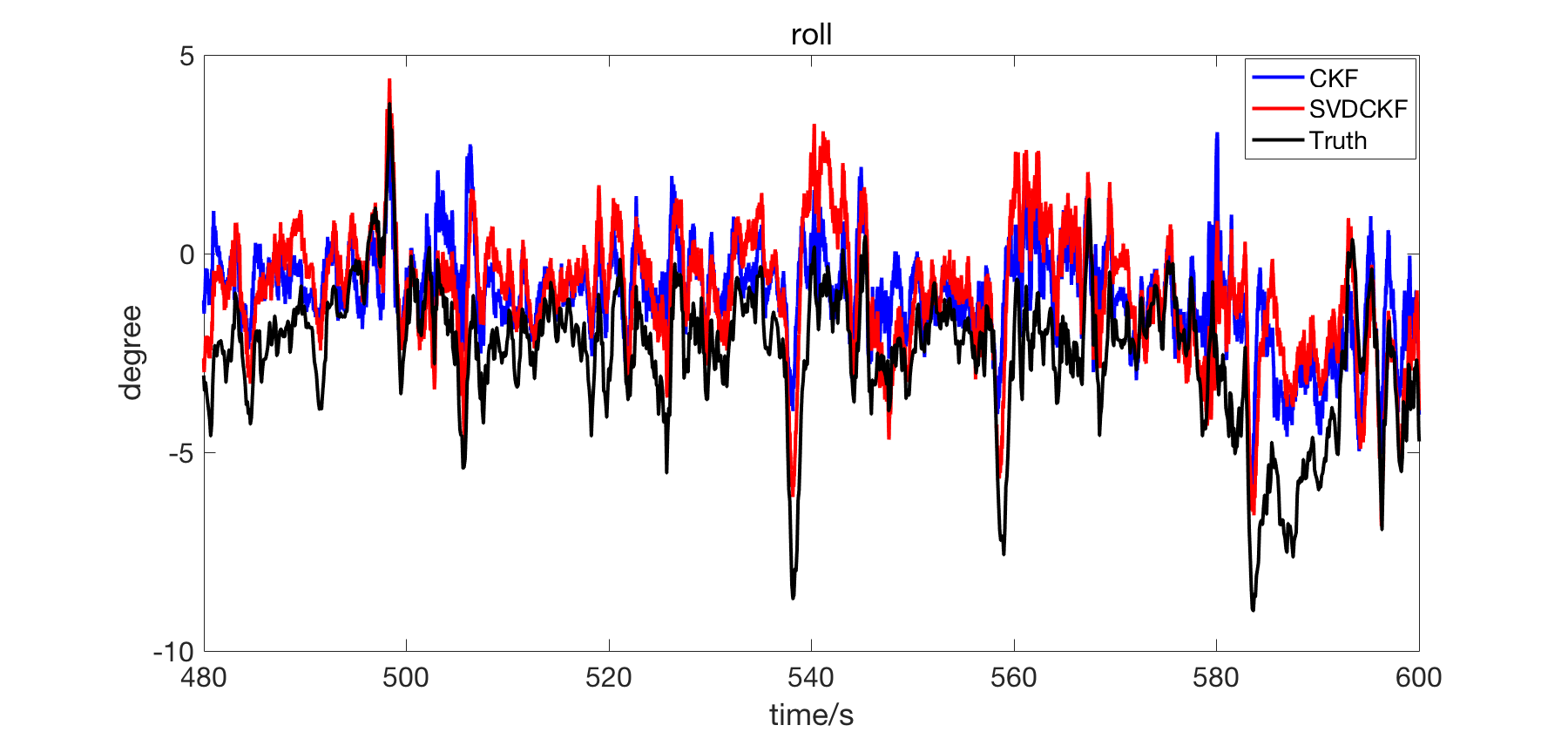} \includegraphics[width=6cm,height=4cm]{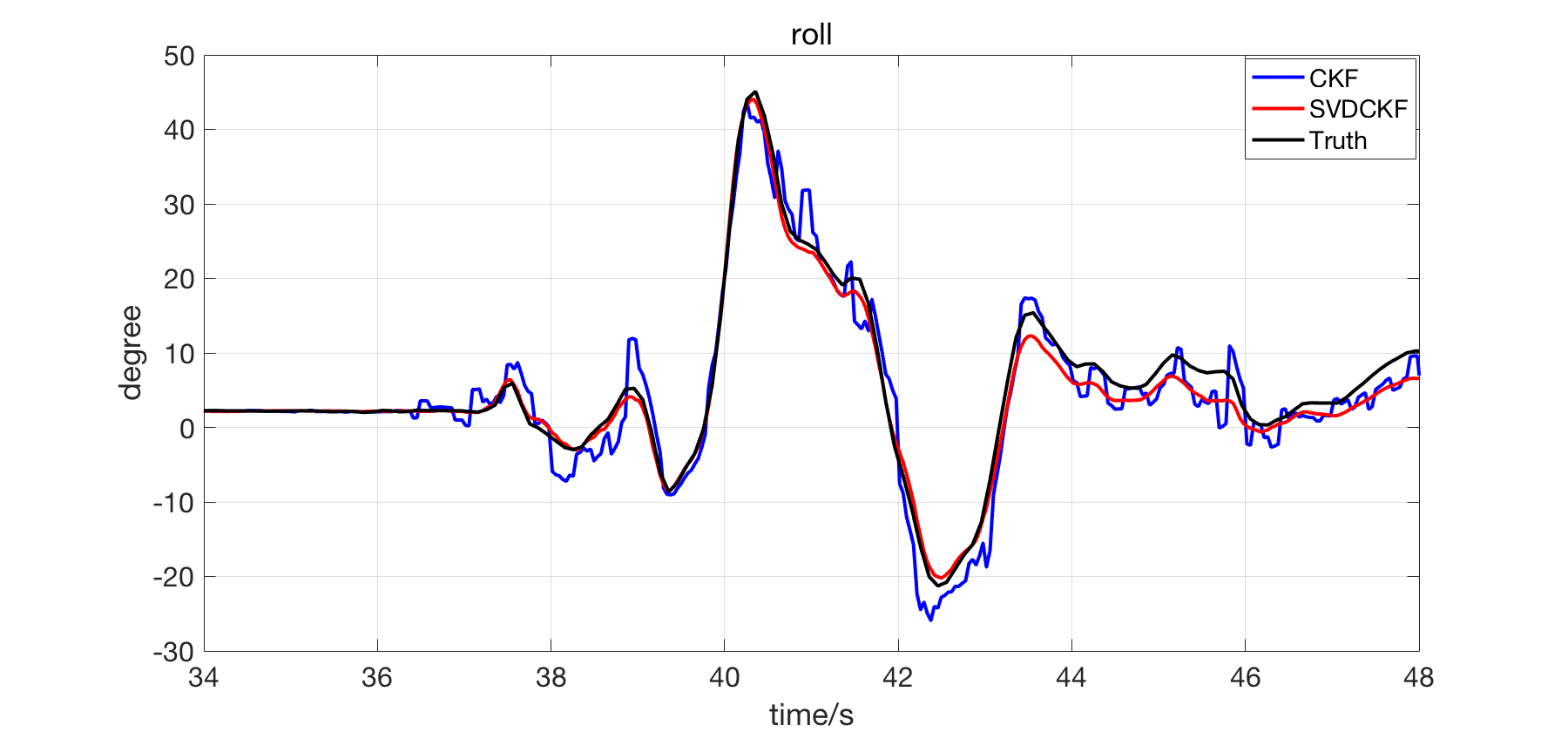}}
 \end{figure}
 \vspace{-13mm}
 \begin{figure}[htbp]
 	 	\centerline{\includegraphics[width=6cm,height=4cm]{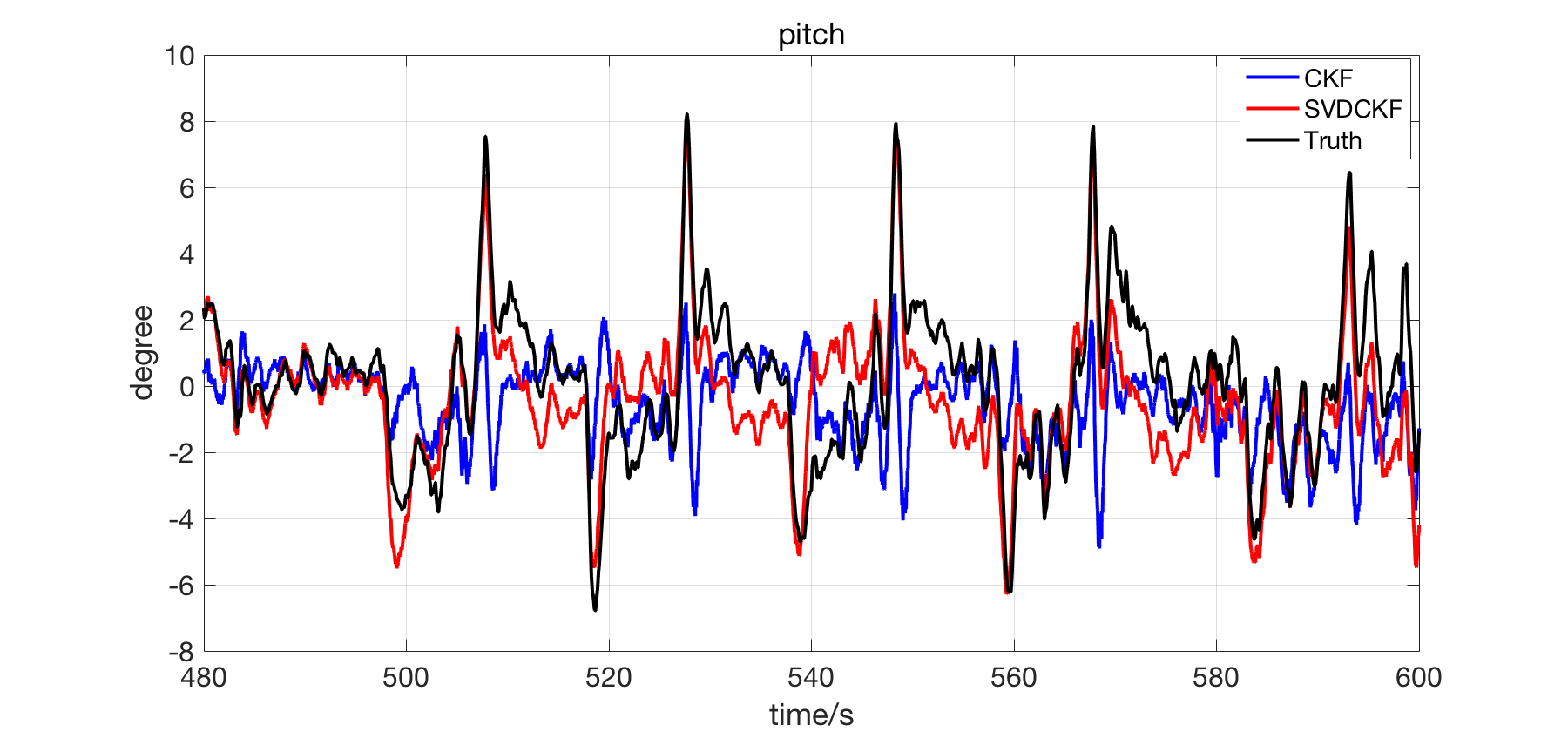} \includegraphics[width=6cm,height=4cm]{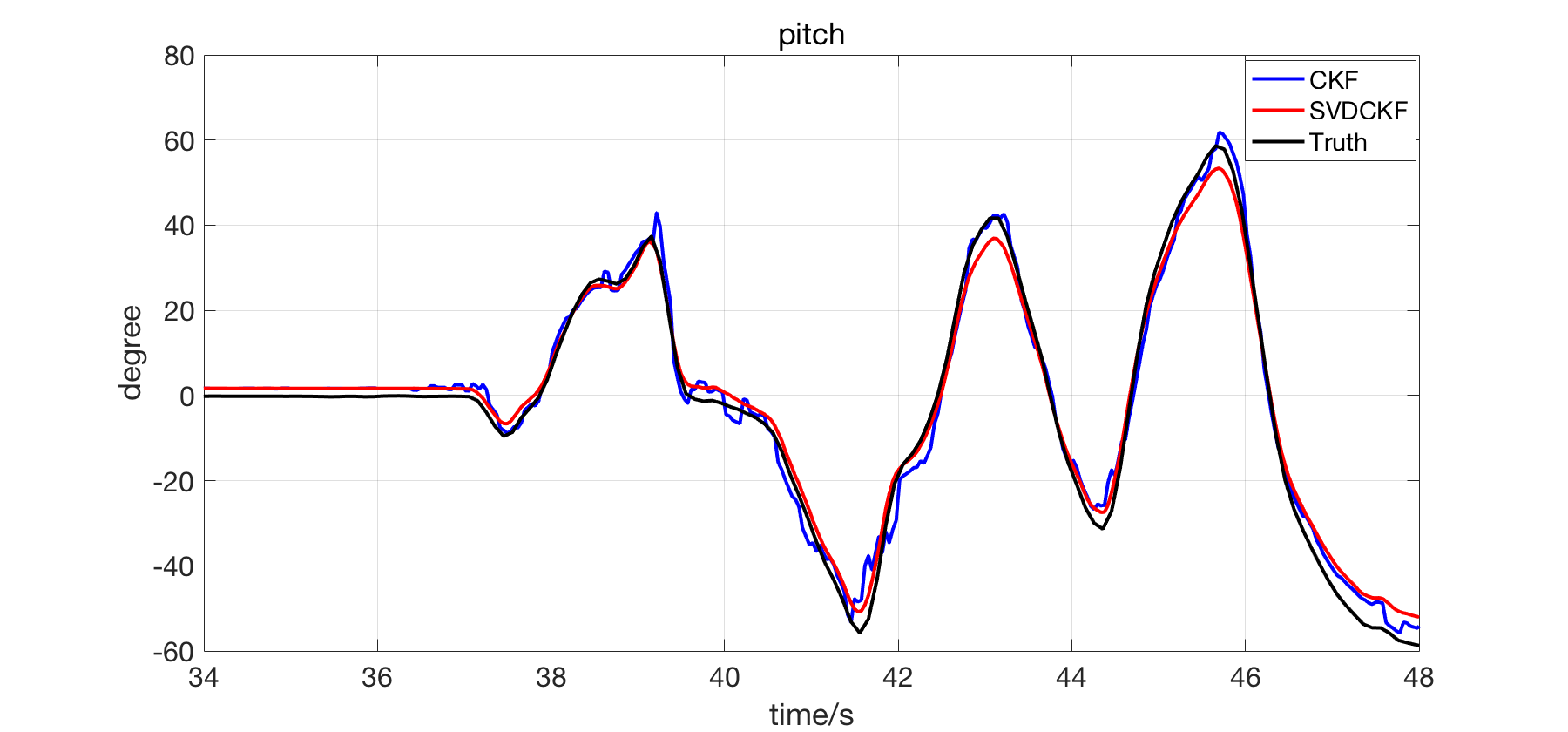}}
 \end{figure}
 \vspace{-10mm}
\begin{figure}[H]
 	\centerline{\includegraphics[width=6cm,height=4cm]{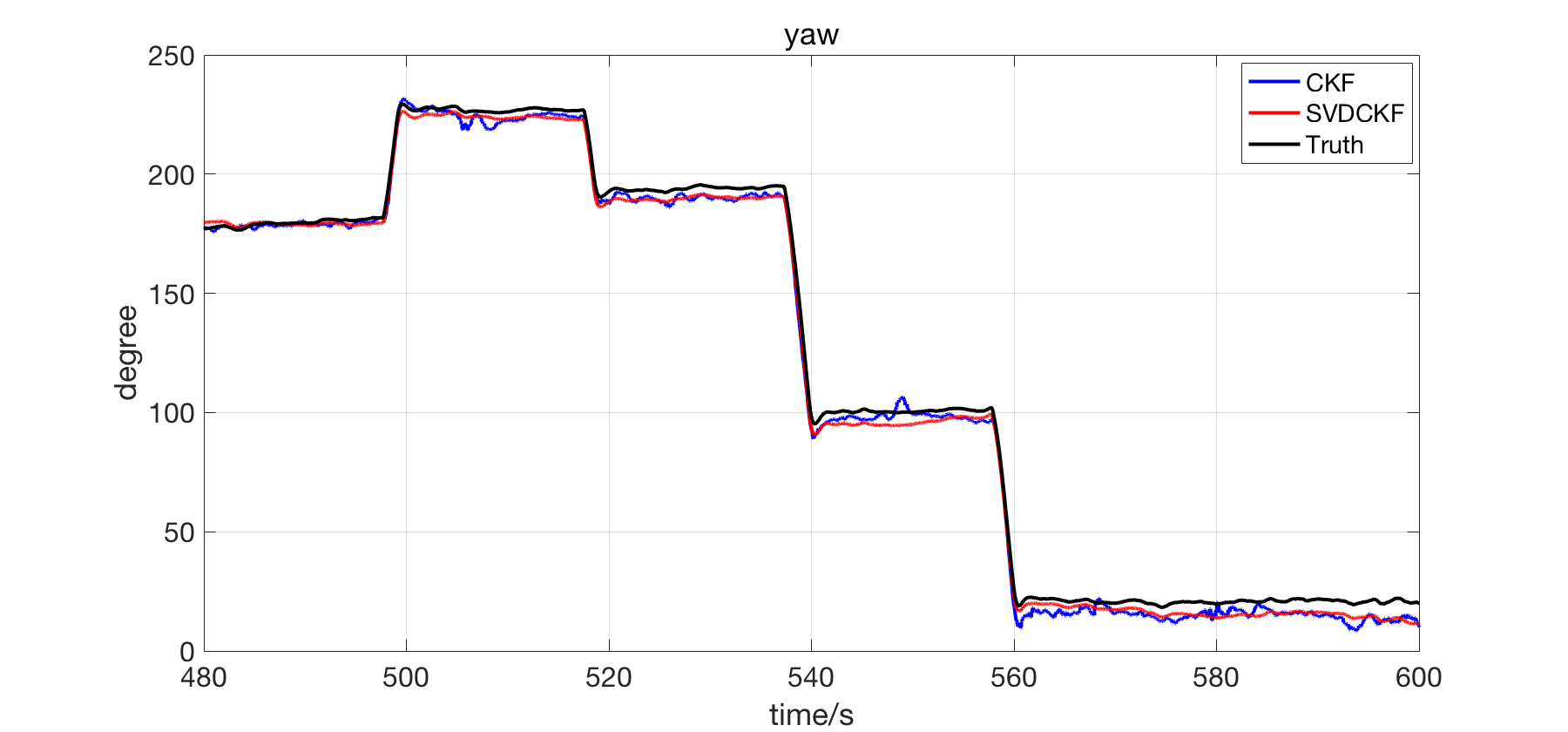} \includegraphics[width=6cm,height=4cm]{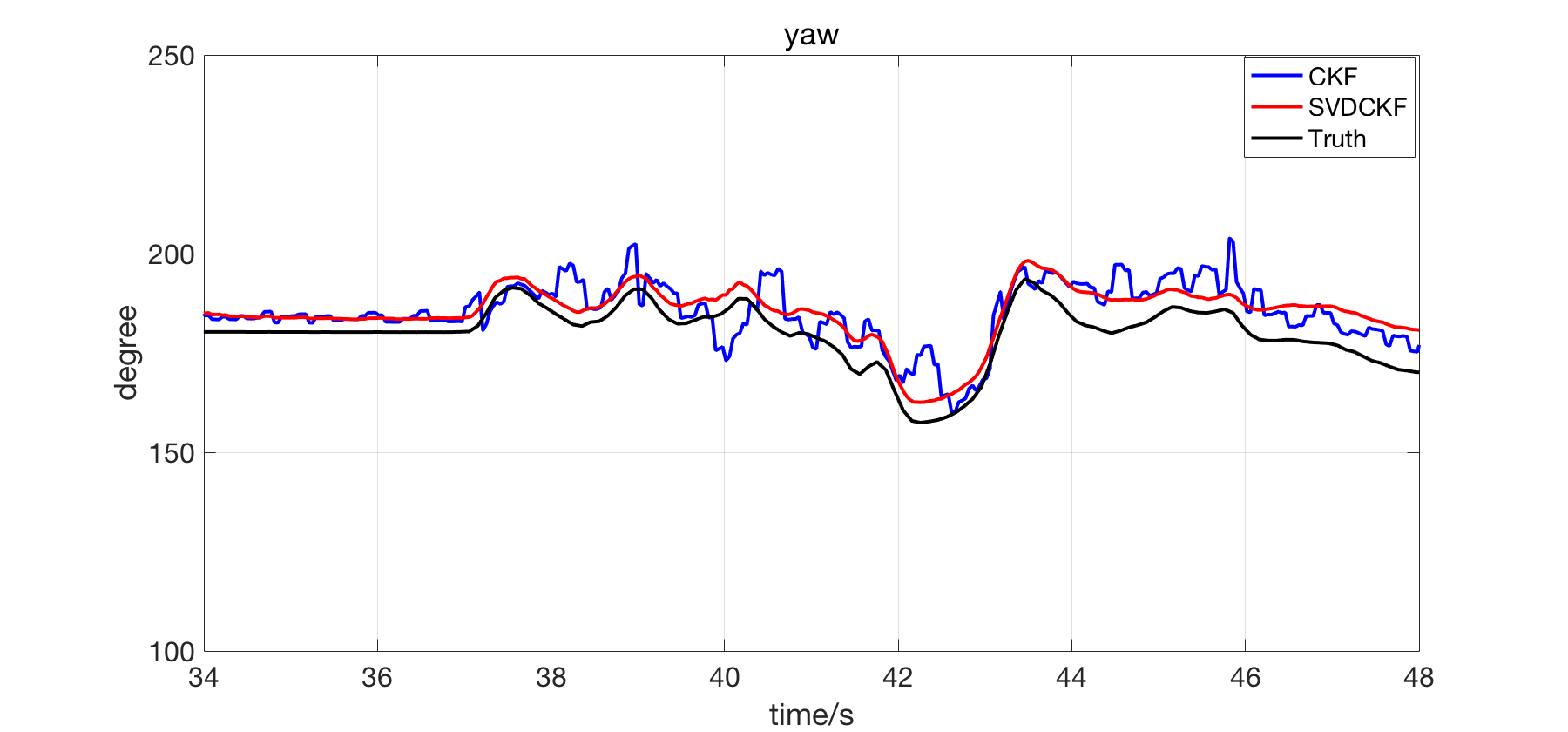}}
 	\caption{ The attitude of low and high dynamic flight conditions} 
 	\label{fig3}
\end{figure}
 \begin{table}[htbp]
 	\caption{Attitude RMSE(/$\deg$) in  low and high dynamic conditions }
 	\begin{tabular}{@{}lllllllllll@{}}
 		\toprule
 		\qquad\qquad low dynamic \\ \midrule
 		Filter           \ \quad\qquad CKF  & SVDCKF   \\ \midrule           
 		$RMSE_{\phi}$        \qquad0.7334   & \quad0.5552   \\ \midrule  
 		$RMSE_{\theta}$      \qquad1.4061   & \quad1.2198   \\ \midrule  
 		$RMSE_{\psi}$        \qquad2.4380   & \quad0.2868   \\ \toprule
 	\end{tabular}
 	\label{Tab1}
 	\qquad\qquad\begin{tabular}{@{}lllllllllll@{}}
 		\toprule
 		\qquad\qquad high dynamic \\ \midrule
 			Filter           \ \quad\qquad CKF  & SVDCKF   \\ \midrule           
 		$RMSE_{\phi}$        \qquad2.8660   & \quad1.7914   \\ \midrule  
 		$RMSE_{\theta}$      \qquad3.7950   & \quad3.6503   \\ \midrule  
 		$RMSE_{\psi}$        \qquad7.4398   & \quad6.0008   \\ \toprule
 	\end{tabular}
 	\label{Tab2}
 \end{table} 

In the Fig.\ref{fig3}, the left half parts are the attitude in the low dynamic flight conditions, compared to the CKF, the attitude solved by the SVDCKF can better follow the Truth. Moreover, the attitude RMSE of SVDCKF is smaller than the CKF, and improve the attitude estimation accuracy. The right half parts are the attitude in the high dynamic flight conditions, the attitude solution curve of CKF occur the some jitter resulting in the flight instability. However, the attitude solution of the SVDCKF is smoother and more robust than the CKF, and has the smaller attitude RMSE.
\section{Conclusion}\label{sec4}
This paper presents an improved nonlinear FastEuler AHRS estimation based on the SVDCKF algorithm for the accurate attitude estimation of the small-UAV. The contributions of this paper are mainly that: (1). the nonlinear quaternion AHRS model and the sensor model are established; (2). The SVDCKF is designed to enhance the filter solution accuracy and solve the non-positive definite of the state covariance matrix $P$; Simulation and experimental results demonstrate that the proposed AHRS filter algorithm can effectively provide the better attitude estimation than SVCDCKF and meets the flying requirements of the small-UAV. 


\begin{thebibliography}{6}
%
\bibitem {Gebre:Demoz:Roger:Powel}
Gebre-Egziabher, Demoz, Roger C. Hayward, and J. David Powell. "A low-cost GPS/inertial attitude heading reference system (AHRS) for general aviation applications." IEEE 1998 Position Location and Navigation Symposium (Cat. No. 98CH36153). IEEE, 1996.
\bibitem {Yadav:Nagesh:Roger:Chris}
Yadav, Nagesh, and Chris Bleakley. "Accurate orientation estimation using AHRS under conditions of magnetic distortion." Sensors 14.11 (2014): 20008-20024.
\bibitem{Welch:Greg:Gary}
Welch, Greg, and Gary Bishop. "An introduction to the Kalman filter." (1995): 41-95.
\bibitem{wang}
Wang, Xiao-Xu, et al. "Overview of deterministic sampling filtering algorithms for nonlinear system." Control and Decision 27.6 (2012): 801-812.
\bibitem{song:Weng}
Song, Yu, Xinwu Weng, and Xingang Guo. "Small UAV Attitude Estimation Based on the Algorithm of Quaternion Extended Kalman Filter." Journal of Jilin University (Science Edition) 53.3 (2015).
\bibitem{Wu}
Yongliang, Wu, et al. "Attitude estimation for small helicopter using extended kalman filter." 2008 IEEE Conference on Robotics, Automation and Mechatronics. IEEE, 2008.
\bibitem{ZHAO}
ZHAO, Lin, et al. "Overview of nonlinear filter methods applied in integrated navigation system [J]." Journal of Chinese Inertial Technology 1 (2009).
\bibitem{wan:Rudolph}
Wan, Eric A., and Rudolph Van Der Merwe. "The unscented Kalman filter for nonlinear estimation." Proceedings of the IEEE 2000 Adaptive Systems for Signal Processing, Communications, and Control Symposium (Cat. No. 00EX373). Ieee, 2000.
\bibitem{Arasaratnam:Simon}
Arasaratnam, Ienkaran, and Simon Haykin. "Cubature kalman filters." IEEE Transactions on automatic control 54.6 (2009): 1254-1269.
\bibitem{Arasaratnam:Sanjeev}
Arulampalam, M. Sanjeev, et al. "A tutorial on particle filters for online nonlinear/non-Gaussian Bayesian tracking." IEEE Transactions on signal processing 50.2 (2002): 174-188.
\bibitem{Pourtakdoust:Ghanbarpour}
Pourtakdoust, S. H., and H. Ghanbarpour Asl. "An adaptive unscented Kalman filter for quaternion‐based orientation estimation in low‐cost AHRS." Aircraft Engineering and Aerospace Technology (2007).
\bibitem{Costanzit:Riccardo}
Costanzi, Riccardo, et al. "An attitude estimation algorithm for mobile robots under unknown magnetic disturbances." IEEE/ASME Transactions on Mechatronics 21.4 (2016): 1900-1911.
\bibitem{Bart:I.:Yaakov}
Bar-Itzhack, I. Y., and Yaakov Oshman. "Attitude determination from vector observations: Quaternion estimation." IEEE Transactions on Aerospace and Electronic Systems 1 (1985): 128-136.
\bibitem{Shepperd:Stanley}
Shepperd, Stanley W. "Quaternion from rotation matrix." Journal of Guidance and Control 1.3 (1978): 223-224.
\bibitem{Higham:Nicholas}
Higham, Nicholas J. Analysis of the Cholesky decomposition of a semi-definite matrix. Oxford University Press, 1990.
\bibitem{Golub:Gene}
Golub, Gene H., and Christian Reinsch. "Singular value decomposition and least squares solutions." Linear Algebra. Springer, Berlin, Heidelberg, 1971. 134-151.
\end{thebibliography}
\end{document}